\documentclass[11pt,a4paper]{article}
\usepackage{cmap}
\usepackage[utf8x]{inputenc}
\usepackage[T1]{fontenc}
\usepackage[english]{babel}
\usepackage{amssymb,amsmath}
\usepackage{amsthm}
\usepackage{bbm}
\usepackage{microtype}
\usepackage{lmodern}

{
\rmfamily
\DeclareFontShape{T1}{lmr}{m}{sc}{<->ssub*cmr/m/sc}{}
\DeclareFontShape{T1}{lmr}{b}{sc}{<->ssub*cmr/b/sc}{}
\DeclareFontShape{T1}{lmr}{bx}{sc}{<->ssub*cmr/bx/sc}{}
}

\newcommand{\thmheadercommand}[1]{\textbf{\scshape{}#1.\\*}}

\newtheoremstyle{yannthm}{\topsep}{\topsep}{\slshape}{}{\scshape\bfseries}{.}{.5em}{%
\thmname{#1}\thmnumber{ #2}\thmnote{#3}%
}
\newtheoremstyle{yannthm2}{\topsep}{\topsep}{}{}{\scshape\bfseries}{.}{.5em}{%
\thmname{#1}\thmnumber{ #2}\thmnote{#3}%
}

\def\R{{\mathbb{R}}}

\renewcommand{\geq}{\geqslant}
\renewcommand{\leq}{\leqslant}

\newcommand{\deq}{\mathrel{\mathop:}=}

\newcommand{\from}{\colon} 

\def\eps{\varepsilon}
\renewcommand{\epsilon}{\varepsilon}
\renewcommand{\phi}{\varphi}

\DeclareMathOperator{\Ent}{Ent}
\DeclareMathOperator{\Var}{Var}

\let\oldPr\Pr
\renewcommand{\Pr}{\oldPr\nolimits}
\newcommand{\E}{\mathbb{E}}
\newcommand{\KL}[2]{\mathrm{KL}\!\left(#1 \,|\hspace{-.15ex}|\,#2\right)}

\DeclareMathOperator{\Tr}{Tr}

\DeclareMathOperator{\Id}{Id}

\newcommand{\abs}[1]{\left|\mskip1mu#1\right|}
\newcommand{\norm}[1]{\left\|#1\right\|}

\newcommand{\1}{\mathbbm{1}}

\newcommand{\twopi}{2\hspace{-.23em}\pi}

\newenvironment{dem}[1][]{\begin{proof}[\thmheadercommand{Proof#1}]~\newline\ignorespaces}{\end{proof}}

{
\theoremstyle{yannthm}
\newtheorem{defi}{Definition}
\newtheorem*{defi*}{Definition}
\newtheorem{prop}[defi]{Proposition}
\newtheorem*{prop*}{Proposition}
\newtheorem{thm}[defi]{Theorem}
\newtheorem*{thm*}{Theorem}
\newtheorem{lem}[defi]{Lemma}
\newtheorem*{lem*}{Lemma}
\newtheorem{cor}[defi]{Corollary}
\newtheorem*{cor*}{Corollary}

\newtheorem*{ex*}{Example}

\newtheorem*{subenonce}{}

\theoremstyle{yannthm2}

\newtheorem*{exo*}{Exercise}

\newtheorem{rem}[defi]{Remark}
\newtheorem*{rem*}{Remark}

\newtheorem*{subenonce2}{}

}

\newcommand{\transp}[1]{#1^{\!\top}\!}

\usepackage[pdfborder={0 0 0}]{hyperref}

\title{Auto-encoders: reconstruction versus compression}
\author{Yann Ollivier}

\newcommand{\Prob}{\mathrm{Prob}}
\newcommand{\D}{\mathcal{D}}

\newcommand{\Lrec}{L_{\mathrm{rec}}}
\newcommand{\Lgen}{L_{\mathrm{gen}}}
\newcommand{\Lfgen}{L_{f\mathrm{\text{-}gen}}}

\newcommand{\LNfgen}{L_{f,\Sigma\mathrm{\text{-}gen}}}
\newcommand{\Ltwopart}{L_{\mathrm{two\text{-}part}}}
\DeclareMathOperator{\diag}{diag}

\newcommand{\empf}{q_f}

\begin{document}
\maketitle

\begin{abstract}
We discuss the similarities and differences between training an
auto-encoder to minimize the reconstruction error, and training the same
auto-encoder to compress the data via a generative model. 
Minimizing a codelength for the data using an auto-encoder is equivalent
to minimizing the reconstruction error plus some correcting terms which
have an interpretation as either a \emph{denoising} or \emph{contractive}
property of the decoding function. These terms are related but not
identical to
those used in
denoising or contractive auto-encoders
\cite{DenoisingAE2010,ContractiveAE2011}. In particular, the codelength
viewpoint fully determines an optimal noise level for the denoising
criterion.
\end{abstract}

Given a dataset, auto-encoders (for instance, \cite[Section 8.1]{HintonPlaut1987}
or \cite{HintonSalakhutdinovScience2006}) aim at building a hopefully simpler
representation of the data via a hidden, usually lower-dimensional
\emph{feature space}. This is done by
looking for a pair of maps $X\stackrel{f}{\to} Y\stackrel{g}{\to} X$ from
data space $X$ to feature space
$Y$ and back, such that the reconstruction error between $x$ and
$g(f(x))$ is small.  Identifying relevant features hopefully makes the
data more understandable, more compact, or simpler to describe.

Here we take this interpretation literally, by considering auto-encoders
in the framework of minimum description length (MDL), i.e., data compression via a
probabilistic generative model, using the general correspondence between
compression and ``simple'' probability distributions on the data
\cite{GrunwaldMDL}. 
The objective is then to minimize the codelength (log-likelihood) of the
data using the features found by the auto-encoder\footnote{The goal here is
not to build an actual compressed code of the data,
but
to find a good pair of feature and generative functions that \emph{would}
yield a short codelength \cite{GrunwaldMDL}. If the codelength is known as a function of the
 parameters of the auto-encoder, it can be used as the training criterion.}.

We use the ``variational'' approach to answer the following
question: Do auto-encoders trained to minimize reconstruction error
actually minimize the length of a compressed encoding of the data, at
least approximately? 

We will see that by adding an information-theoretic term to the
reconstruction error, auto-encoders can be trained to minimize a tight
upper bound on the codelength (compressed size) of the data.

In Section~\ref{sec:twopart} we introduce a first, simple bound on
codelength based on reconstruction error: a dataset $\D\subset X$ can be
encoded by encoding a (hopefully
simpler) feature value $f(x)$ for each $x\in \D$, and applying the decoding
function $g$. However, this result only applies to discrete features, and the
resulting bound is far from tight.
Still, this already illustrates how minimizing
codelength favors using fewer features.

In Section~\ref{sec:main} we refine the bound from
Section~\ref{sec:twopart} and make it valid for general feature spaces
(Proposition~\ref{prop:main}).
This bound is tight in the sense that it gets arbitrarily
close to the actual codelength when the feature and generative functions
are inverse to each other in a probabilistic sense. This is an instance
of the \emph{variational bound} \cite[Chapter 10]{Bishop_book}.  A related result appears in \cite[Section 2.2]{VariationalAE}.

The result in Section~\ref{sec:main} also illustrates how, to optimize codelength, an auto-encoder
approach helps compared to directly looking for a generative model.
Trying to optimize the codelength directly is often difficult
(Section~\ref{sec:grad}). So even though the codelength $\Lgen$ depends
only on the generative function $g$ and not on a feature function, we
build an upper bound on $\Lgen$ depending on both; optimizing over $g$
aims at lowering $\Lgen$ by lowering this upper bound, while optimizing
over $f$ aims at making the upper bound more precise.

In Sections~\ref{sec:continuous} and~\ref{sec:optnoise} we provide a
connection with \emph{denoising
auto-encoders}
\cite{DenoisingAE2010}. When the feature space is continuous, it is
impossible to encode a feature value $f(x)$ exactly for each $x$ in the
dataset as this yields an infinite codelength. Thus, it is necessary to
encode features with finite precision and to use a decoding function that
is not too sensitive to approximate features. Quantifying this effect
leads to an explicit upper bound on codelength (Corollary~\ref{cor:DAE}).
The denoising criterion is from
features to output, rather than from input to features as in
\cite{DenoisingAE2010}.

Moreover the MDL approach allows us to find the optimal noise level for
the denoising criterion, i.e., the one which yields the best codelength
(Theorem~\ref{thm:optnoise}).  In particular, the noise level should
be set differently for each data sample.

In Section~\ref{sec:contract} we establish
a connection with \emph{contractive auto-encoders}
\cite{ContractiveAE2011}: under various approximations, minimizing
codelength penalizes large derivatives of the output
(Proposition~\ref{prop:CAE}). The
penalty takes a form somewhat different  from \cite{ContractiveAE2011},
though: contractivity occurs from features to output rather than from
input to features,
and the penalty term is not the Frobenius norm of the
Jacobian matrix but the sum of the logs of the norms of its rows. An
advantage of the MDL approach is that the penalty constant is determined
from theory.

In Section~\ref{sec:variance} we show that optimal compression requires
including the variance of each data component as additional parameters,
especially when various data components have different variances or noise
levels.  Compression focuses on relative rather than absolute error,
minimizing the \emph{logarithms} of the errors.

The variational bound has already been applied to neural networks in
non-auto-encoding situations, to evaluate the cost of encoding the
network parameters \cite{Graves_varinf,HintonCamp_varinf}. In that
situation, one tries to find a map $Y\stackrel{g}{\to} X$ that minimizes the
codelength of the output data $x$ \emph{if the features $y$ are given}; this
decomposes as the output error plus a term describing the cost of
encoding the parameters of $g$.
In an auto-encoding setting $X\stackrel{f}{\to} Y\stackrel{g}{\to} X$, it
is meaningless to encode the dataset given the very same inputs: so the
dataset is encoded by
encoding the features $y$ together with $g$. In this text we focus on the
cost of encoding $y$, and the consequences of minimizing the resulting
codelength. Encoding of the parameters of $g$ can be done 
following \cite{Graves_varinf} and we do not reproduce it here.
Still, the cost of $g$ must be included for actual data compression, and
also especially when comparing generative models with different
dimensions.

\paragraph{Notation: Auto-encoders, reconstruction error.}
Let $X$ be an input space and $Y$ be a feature
space, usually of smaller dimension. $Y$ may be discrete, such as $Y=\{0,1\}^d$ (each feature present/absent) or
$Y=\{1,\ldots,d\}$ (classification), or continuous.

An auto-encoder can be seen as a pair of functions $f$ and $g$, the
\emph{feature} function and the \emph{generative} function. The feature
function goes from $X$ to $Y$ (deterministic features) or to $\Prob(Y)$
(probability distribution on features), while the generative function
goes from $Y$ to $X$ or $\Prob(X)$.

The functions $f$ and $g$ depend on parameters $\theta_f$ and $\theta_g$
respectively. For instance, $f$ and $g$ may each represent a multilayer
neural network or any other model.  Training the parameters via the reconstruction error
criterion focuses on having $g(f(x))$ close to $x$, as follows.

Given a feature function $f\from X\to Y$ and a generative function
$g\from Y\to
\Prob(X)$, define the \emph{reconstruction error} for a dataset
$\D\subset X$ as
\begin{equation}
\Lrec(x)\deq -\log g_{f(x)}(x),
\qquad
\Lrec(\D)\deq \sum_{x\in\D} \Lrec(x)
\end{equation}
where $g_y$ is the probability distribution on $X$ associated with
feature $y$.

The case of a deterministic $g\from Y \to X$ with square error
$\norm{g(f(x))-x}^2$ is recovered by interpreting $g$ as a Gaussian
distribution\footnote{While the choice of variance does not influence
minimization of the reconstruction error, when working with codelengths it
will change the scaling of the various terms in
Propositions~\ref{prop:twopart}--\ref{prop:CAE}. See Section~\ref{sec:variance} for the
optimal variance}
centered at $g(f(x))$. So we will always consider that $g$ is
a probability distribution on $X$.

Discrete-valued features can be difficult to train using gradient-based
methods. For this reason, with discrete features it is more natural to
define $f(x)$ as a distribution over the feature space $Y$ describing the
law of inferred features for $x$.
Thus $f(x)$ will have
continuous parameters.
If $f\from X\to \Prob(Y)$ describes a probability distribution on
features for
each $x$, we define the expected reconstruction error as the expectation of the above:
\begin{equation}
\E\Lrec(x)\deq -\E_{y\sim f(x)} \log g_y(x),\qquad
\E\Lrec(\D)\deq \sum_{x\in \D} \E\Lrec(x)
\end{equation}
This covers the previous case when $f(x)$ is a Dirac mass at a
single value $y$.

{\small In Sections~\ref{sec:aagen}--\ref{sec:main} the logarithms may be
in any base; in Sections~\ref{sec:continuous}--\ref{sec:variance} the
logarithms are in base $\mathrm{e}$.}

\paragraph{Auto-encoders as generative models.}\label{sec:aagen} Alternatively, auto-encoders can be viewed as generative models for the
data. For this we assume that we are given (or learn) an elementary model
$\rho$ on feature space, such as a Gaussian or Bernoulli model, or even a
uniform model in which each feature is present or absent with
probability $1/2$. Then, to generate the data, we draw features at random
according to $\rho$ and apply the generative function $g$. The goal is to
maximize the probability to generate the actual data. In this viewpoint
the feature function $f$ is used only as a prop to learn a good feature
space and a good generative function $g$.

Given a probability distribution $p$ on a set $X$, a dataset
$(x_1,\ldots,x_n)$ of points on $X$ can be encoded in $-\sum_i \log_2
p(x_i)$ bits\footnote{Technically, for continuous-valued data $x$, the
actual compressed length is
rather $-\log_2 p(x)-\log_2\eps$ where $\eps$ is the quantization
threshold of the data and $p$ is the probability density for $x$. For the
purpose of comparing two different probabilistic models $p$ on the same
data with the same $\eps$, the term $-\log_2\eps$ can be dropped}. Let $\rho\in\Prob(Y)$ be the elementary model on feature
space and let $g\from Y\to \Prob(X)$ be the generative function. The
probability to obtain $x\in X$ by drawing $y\sim \rho$ and applying $g$ is
\begin{equation}
p_g(x)\deq \int_y \rho(y) g_y(x)
\end{equation}
(where the integral is a sum if the feature space $Y$ is discrete). Thus
minimizing the codelength of the dataset $\D$ amounts to minimizing
\begin{align}
\Lgen(\D)&\deq \sum_{x\in\D}\Lgen(x),
\\
\Lgen(x)&\deq  -\log p_g(x)=-\log \int_y \rho(y)g_y(x)
\end{align}
over $g$.

This is the codelength of the data knowing the distribution $\rho$ and
the function $g$. We do not consider here the problem of encoding the
parameters of $\rho$ and $g$; this can be done following
\cite{Graves_varinf}, for instance.

\label{sec:grad}
The codelength $\Lgen$ does not depend on any feature function $f$.
However, it is difficult to optimize $\Lgen$ via a direct approach: this leads to working with all possible
values of $y$ for every sample $x$, as $\Lgen(x)$ is an integral over
$y$. Presumably, for each given $x$ only a few feature values contribute
significantly to $\Lgen(x)$.
Using a feature function is a way to explore fewer possible
values of $y$ for a given $x$, hopefully those that contribute most to
$\Lgen(x)$.

For instance, consider the gradient of $\Lgen(x)$
with respect to a parameter $\theta$:
\begin{align}
\frac{\partial \Lgen(x)}{\partial \theta}&=-\frac{\int_y \rho(y) \partial
g_y(x)/\partial \theta}{\int_y \rho(y) g_y(x)}
=-\frac{\int_y \rho(y) g_y(x)\partial \ln
g_y(x)/\partial \theta}{\int_y \rho(y) g_y(x)}
\\&=
-\E_{y\sim p_g(y|x)}\frac{\partial \ln
g_y(x)}{\partial \theta}
\end{align}
where $p_g(y|x)=\rho(y) g_y(x)\,/\int_{y'} \rho(y') g_{y'}(x)$ is the
conditional probability of $y$ knowing $x$, in the generative model given
by $\rho$ and $g$. In general we have no easy access to this
distribution.

Using a (probabilistic) feature function $f$ and
minimizing the reconstruction error $\E\Lrec(x)$ amounts to replacing the
expectation under $y\sim p_g(y|x)$ with an expectation under $f(x)$ in
the above, presumably easier to handle. However this gives no guarantees about
minimizing $\Lgen$ unless we know that the feature function $f$ is close
to the inverse of the generative function $g$, in the sense that
$f(x)(y)$ is close to the conditional
distribution $p_g(y|x)$ of $y$ knowing $x$. It
would be nice to obtain a guarantee on the codelength based on the
reconstruction error of a feature function $f$ and
generative function $g$.

The variational bound in Proposition~\ref{prop:main} below shows that,
given a feature function $f$ and a generative function $g$, the quantity
$\Lrec(x)+\KL{f(x)}{\rho}$ is an upper bound on the codelength
$\Lgen(x)$. Training an autoencoder to minimize this criterion will thus
minimize an upper bound on $\Lgen$.

Moreover,
Proposition~\ref{prop:main} shows that the bound is tight when $f(x)$ is
close to $p_g(y|x)$, and that minimizing this bound will indeed
bring $f(x)$ closer to $p_g(y|x)$.  On the other hand, just
minimizing the reconstruction error does not, a priori, guarantee any of
this.

\paragraph{Two-part codes: explicitly encoding feature values.}
\label{sec:twopart} We first
discuss a simple, less efficient ``two-part'' \cite{GrunwaldMDL} coding method. It always yields a
codelength larger than $\Lgen$ but is more obviously related to the
auto-encoder reconstruction error. 

Given a generative model $g\from Y\to \Prob(X)$ and a prior\footnote{We
use the term ``prior'' in a loose way: 
we just encode features $y$ with a code of length
$-\log \rho(y)$, without implying any a priori belief. Thus $\rho$ is
just a simple model used on feature space.} distribution
$\rho$ on $Y$, one way to encode a data sample $x\in X$ is to explicitly
encode a well-chosen feature value $y\in Y$ using the prior distribution $\rho$ on
features, then encode $x$ using the
probability distribution $g_y(x)$ on $x$ defined by $y$. The
codelength resulting from this choice of $y$ is thus
\begin{equation}
\label{eq:Ltwopart}
\Ltwopart(x)\deq -\log \rho(y)-\log g_y(x)
\end{equation}

In this section we assume that $Y$ is a discrete set. Indeed for
continuous features, the above does not make sense as encoding a precise
value for $y$ would require an infinite codelength. Continuous features
are dealt with in Sections~\ref{sec:main} and~\ref{sec:continuous}.

We always have 
\begin{equation}
\Ltwopart(x)\geq \Lgen(x)
\end{equation}
for discrete features, as $\Ltwopart$ uses a single
value of $y$ while $\Lgen$ uses a sum over $y$. The
difference can be substantial if, for instance, not all feature
components are relevant for all $x$: using the two-part code, it is
always necessary to fully encode the feature values $y$.

From an auto-encoder perspective, the feature function $f$ is used to
choose the feature value $y$ used to encode $x$. So if the feature
function is deterministic, $f\from X\to Y$, and if we set $y=f(x)$ in the
above, the cost of encoding the dataset
is
\begin{align*}
\Ltwopart(\D)&=-\sum_{x\in\D} \left(\log \rho(f(x))+\log g_{f(x)}(x)\right)
\\&=\Lrec(\D)-\sum_{x\in\D} \log\rho(f(x))
\end{align*}
involving the reconstruction error and a cross-entropy term between the
empirical distribution of features $f(x)$ and the prior $\rho$ on feature
space. We can further decompose
\begin{equation}
-\frac{1}{\#\D}\sum_{x\in\D} \log\rho(f(x))=
\KL{\empf}{\rho}+\Ent \empf
\end{equation}
where $\empf$ is the empirical distribution of the feature $f(x)$ when
$x$ runs over the dataset,
\begin{equation}
\empf(y)\deq \frac{1}{\#\D}\sum_{x\in\D} \1_{f(x)=y}
\end{equation}
and $\KL{q_f}{\rho}=\E_{y\sim q_f}\log (q_f(y)/\rho(y))$ is the Kullback--Leibler
divergence between $q_f$ and $\rho$.

If the feature function $f$ is probabilistic, $f\from X\to\Prob(Y)$, the
analysis is identical, with the expected two-part codelength of $x$ being
\begin{align}
\E\Ltwopart(x)&=\E_{y\sim f(x)} (-\log \rho(y)-\log
g_y(x))\\&=\E\Lrec(x)-\E_{y\sim f(x)} \log \rho(y)
\end{align}

Thus we have proved the following, which covers both the case of
probabilistic $f$ and of deterministic $f$ (by specializing $f$ to a
Dirac mass) on a discrete feature space.

\begin{prop}[ (Two-part codelength and reconstruction error for discrete
features)]
\label{prop:twopart}
The expected two-part codelength of $x\in \D$ and the reconstruction
error are related by
\begin{align}
\label{eq:twopartmain}
\E\Ltwopart(\D)&=\E\Lrec(\D)-\sum_{x\in\D} \E_{y\sim f(x)} \log
\rho(y)
\\&=\E\Lrec(\D)+(\#\D)(\KL{\empf}{\rho}+\Ent \empf)
\end{align}
where
\begin{equation}
\empf(y)\deq \frac{1}{\#\D}\sum_{x\in\D} \Pr(f(x)=y)
\end{equation}
is the empirical distribution of features.
\end{prop}

Here are a few comments on this relation. These comments also apply to
the codelength discussed in Section~\ref{sec:main}.

\begin{itemize}
\item The reconstruction error in~\eqref{eq:twopartmain} is the
\emph{average} reconstruction error for features $y$ sampled from $f(x)$,
in case $f(x)$ is probabilistic. For instance, applying
Proposition~\ref{prop:twopart} to neural
networks requires interpreting the activities of the $Y$
layer as probabilities to sample $0/1$-valued features on the $Y$
layer. (This is not necessary for the results of
Sections~\ref{sec:main}--\ref{sec:contract}, which hold for continuous
features.)

\item The cross-entropy term $-\E_{y\sim \empf} \log
\rho(y)=\KL{\empf}{\rho}+\Ent \empf$ is an added term
to the optimisation problem. The Kullback--Leibler divergence favors
feature functions that do actually match an elementary model on $Y$,
e.g., feature distributions that are ``as Bernoulli-like'' as possible. The
entropy term $\Ent \empf$ favors parsimonious feature functions that use fewer
feature components if possible, arguably introducing some regularization or
sparsity.
(Note the absence of any arbitrary parameter in front of this
regularization term:
its value is fixed by the MDL interpretation.)

\item If the elementary model $\rho$ has tunable parameters (e.g., a
Bernoulli parameter for each feature), these come into the optimization
problem as well. If $\rho$ is elementary it will be fairly easy to tune
the parameters to find the elementary model $\rho^\ast(f)$ minimizing the Kullback--Leibler divergence to $\empf$.
Thus in this case the optimization problem over $f$ and $g$ involves a term
$\KL{q_f}{\rho^\ast(f)}$ between the empirical distribution of features and the
closest elementary model.

\end{itemize}

This two-part code is somewhat naive in case not all feature components
are relevant for all samples $x$: indeed for every $x$, a value of $y$
has to be fully encoded. For instance, with feature space $Y=\{0,1\}^d$,
if two values of $y$ differ in one place and contribute equally to
generating some sample $x$, one could expect to save one bit on the
codelength, by leaving a blank in the encoding where the two values of
$y$ differ. In general, one could expect to save $\Ent f(x)$ bits on the
encoding of $y$ if several $y\sim f(x)$ have a high probability to
generate $x$. We now show that indeed $\E\Ltwopart(x)-\Ent f(x)$ is still
an upper bound on $\Lgen(x)$.

\paragraph{Comparing $\Lgen$ and $\Lrec$.} \label{sec:main}
We now turn to the actual
codelength $\Lgen(x)=-\log p_g(x)$ associated with the probabilistic model
$p_g(x)$ defined by the generative function $g$ and the prior $\rho$ on
feature space. As mentioned above, it is always smaller that the two-part
codelength.

Recall that this model first picks a feature value $y$ 
at random according
to the distribution $\rho$ and then generates an object $x$ according to
the distribution $g_y(x)$, so that the associated codelength is $-\log
p_g(x)=-\log \int_y \rho(y) g_y(x)$.

So far this does not depend on the feature function so it is not clear
how $f$ can help in optimizing this codelength. Actually each choice of $f$
leads to upper bounds on $\Lgen$: the two-part codelength
$\Ltwopart$ above is one such bound in the discrete case, and
we now introduce a more precise and more general
one, $\Lfgen$.

We have argued above (Section~\ref{sec:grad}) that for gradient-based
training it would be helpful to be able to sample features from the
distribution $p_g(y|x)$, and it is natural to expect the feature function
$f(x)$ to approximate $p_g(y|x)$, so that $f$ and $g$ are inverse to each
other in a probabilistic sense. The tightness of the bound $\Lfgen$ is
related to the quality of this approximation.  Moreover, while
auto-encoder training based on the reconstruction error provides no
guarantee that $f$ will get closer to $p_g(y|x)$, minimizing $\Lfgen$
does.

\begin{prop}[ (Codelength and reconstruction error for probabilistic
features)]
\label{prop:main}
The codelength $\Lgen$ and reconstruction error $\Lrec$ for an auto-encoder with feature
function $f\from X\to \Prob(Y)$ and generative function $g\from Y\to\Prob(X)$
satisfy
\begin{align}
\label{eq:main}
\Lgen(x)&=\E\Lrec(x)+\KL{f(x)}{\rho}-\KL{f(x)}{p_g(y|x)},
\\
\label{eq:mainD}
\Lgen(\D)&=\E\Lrec(\D)+\sum_{x\in
\D}\KL{f(x)}{\rho}-\sum_{x\in\D} \KL{f(x)}{p_g(y|x)}
\end{align}
where $\rho$ is the elementary model on features, and
$p_g(y|x)=\frac{\rho(y)g_y(x)}{\int_{y'} \rho(y')g_{y'}(x)}$.

In particular, for any feature function $f$, the quantity
\begin{equation}
\Lfgen(\D)\deq \sum_{x\in\D} \Lfgen(x)
\end{equation}
where
\begin{equation}
\Lfgen(x)\deq
\E\Lrec(x)+\KL{f(x)}{\rho}
\end{equation}
is an upper bound on the codelength $\Lgen(\D)$ of the generative function
$g$.
\end{prop}

The result
holds whether $Y$ is discrete or continuous.

The proof is by substitution in the right-hand-side
of~\eqref{eq:main}; actually this is an instance of the \emph{variational
bound} \cite[Chapter 10]{Bishop_book}. A closely related result appears
in \cite[Section 2.2]{VariationalAE}.

On a discrete feature space, $\Lfgen$ is always smaller than the
codelength $\Ltwopart$ above; indeed
\begin{equation}
\Lfgen(x)=\E\Ltwopart(x)-\Ent f(x)
\end{equation}
as can be checked directly.

The term $\KL{f(x)}{\rho}$ represents the cost of encoding a feature
value $y$ drawn from $f(x)$ for each $x$ (encoded using the distribution
$\rho$).  The last,
negative term in~\eqref{eq:main}--\eqref{eq:mainD} represents how pessimistic the reconstruction error is
w.r.t.\ the true codelength when $f(x)$ is far from the feature values that
contribute most to $\Lgen(x)$.

The codelength $\Lgen$ depends only on $g$ and not on the feature
function $f$, so that the right-hand-side
in~\eqref{eq:main}--\eqref{eq:mainD} is the same for all $f$ despite
appearances. Ideally,
this relation could be used to evaluate $\Lgen(\D)$ for a given
generative function $g$, and then to minimize this quantity over $g$.
However, as explained above, the conditional probabilities $p_g(y|x)$ are
not easy to work with, hence the introduction of the upper bound
$\Lfgen$, which does depend on the feature function $f$.

Minimizing $\Lfgen$ over $f$ will bring $\Lfgen$ closer to $\Lgen$.
Since $\Lgen(\D)=\Lfgen(\D)-\sum_{x\in\D} \KL{f(x)}{p_g(y|x)}$, and since
$\Lgen$ does not depend on $f$,
minimizing $\Lfgen$ is the same as bringing
$f(x)$ closer to $p_g(y|x)$ on average.
Thus, in the end, an auto-encoder 
trained by minimizing $\Lfgen$ as a function of $f$ and $g$ will both
minimize an upper bound on the codelength $\Lgen$ and bring $f(x)$ close
to the ``inverse'' of $g$.

This also clarifies the role of the auto-encoder structure in minimizing
the codelength, which does not depend on a feature function: Optimizing
over $g$ aims at actually reducing the codelength by decreasing an upper
bound on it, while optimizing over $f$ will make this upper bound more
precise.

%
%
%

One can apply to $\Lfgen$ the same decomposition as for the two-part codelength, and
write
\begin{equation}
\Lfgen(\D)=\E\Lrec(\D)+(\#\D)(\KL{q_f}{\rho}+\Ent \empf) -\sum_{x\in \D} \Ent f(x)
\end{equation}
where as above $\empf=\frac{1}{\#\D}\sum_{x\in \D} f(x)$ is the empirical feature
distribution. As above, the term $\KL{q_f}{\rho}$ favors feature
distributions that match a simple model. The terms $\Ent \empf$ and
$\sum_{x\in\D} \Ent f(x)$ pull in different directions. Minimizing $\Ent \empf$
favors using fewer features overall (more compact representation).
Increasing the entropy of $f(x)$ for a given $x$, if it can be done
without impacting the reconstruction error, means that more features are
``indifferent'' for reconstructing $x$ and do not have to be encoded, as
discussed at the end of Section~\ref{sec:twopart}.

The ``auto-encoder approximation'' $x'=x$ from \cite[Section
2.4]{deeptrain} can be used to define another bound on $\Lgen$, but is
not tight when $f(x)\approx p_g(y|x)$.

\paragraph{Continuous-valued features and denoising.}\label{sec:continuous}
Proposition~\ref{prop:main} cannot be directly applied to a
deterministic feature function $f\from X\to Y$ with values in a continuous
space $Y$. In the continous case, the reconstruction error based on a
single value $y\in Y$ cannot control the codelength $\Lgen(x)$, which
involves an integral over $y$.
In the setting of Proposition~\ref{prop:main}, a deterministic $f$ seen as a probability
distribution is a Dirac mass at a single value, so that the term
$\KL{f(x)}{\rho}$ is infinite: it is infinitely costly to encode the
feature value $f(x)$ exactly.


This can be overcome by considering the feature values $y$ as probability
distributions over an underlying space $Z$, namely, $Y\subset \Prob(Z)$.
Then Proposition~\ref{prop:main} can be applied to $f(x)$ seen as a
probability distribution over the feature space $Z$.

For instance, one possibility for neural networks with logistic
activation function is to see the activities $y\in [0;1]$ of the feature
layer as
Bernoulli probabilities over discrete-valued binary features, $Z=\{0,1\}$.

One may also use Gaussian distributions over $Z=Y$ and apply
Proposition~\ref{prop:main} to a normal distribution
$\mathcal{N}(f(x),\Sigma)$ centered at $f(x)$ with small covariance
matrix $\Sigma$. Intuitively we overcome the problem of infinite
codelength for $f(x)$ by encoding $f(x)$ with finite accuracy given by
$\Sigma$. 

The reconstruction error $\Lrec$ from Proposition~\ref{prop:main} then
becomes an expectation over features sampled around $f(x)$: this is
similar to \emph{denoising} auto-encoders \cite{DenoisingAE2010}, except that here
the noise is added to the features rather than the inputs. This
relationship is not specific to a particular choice of feature noise
(Bernoulli, Gaussian...) but leads to interesting developments in the
Gaussian case, as follows.

\begin{cor}[ (Codelength and denoising the features)]
\label{cor:DAE}
Let $f\from X\to Y$ be a deterministic feature function with values in
$Y=\R^d$. Let $\Sigma$ be any positive definite matrix. Then
\begin{equation}
\label{eq:gaussianbound}
\Lgen(x)\leq \E\Lrec(x) - \E_{y\sim \mathcal{N}(f(x),\Sigma)} \log \rho(y)
-\frac12\log\det \Sigma
-\frac{d}{2}(1+\log \twopi)
\end{equation}
where $\E\Lrec(x)$ is the expected reconstruction error obtained from a
feature $y\sim \mathcal{N}(f(x),\Sigma)$.

If the elementary model $\rho$ on feature space is
$\mathcal{N}(0,\lambda\Id)$ this reads
\begin{equation}
\Lgen(x)\leq \E\Lrec(x) +
\frac{\norm{f(x)}^2}{2\lambda}+\frac{1}{2\lambda}\Tr(\Sigma)-\frac12\log\det\Sigma+\frac{d}{2}\log
\lambda -\frac{d}{2}
\end{equation}
\end{cor}

Thus the codelength bound decomposes as the sum of the average (noisy)
reconstruction error, constant terms, and a term that penalizes
improbable feature values under the elementary model.

\begin{dem}
Apply Proposition~\ref{prop:main} with a normal distribution
$\mathcal{N}(f(x),\Sigma)$ as the feature distribution.
\end{dem}

We refer to \cite[Section 4.2]{DenoisingAE2010} for a discussion and
further references on training with noise in an auto-encoder setting.

In practice, the bound~\eqref{eq:gaussianbound} can be optimized over $f$ and $g$ via Monte Carlo sampling
over $y\sim \mathcal{N}(f(x),\Sigma)$.
For the case of neural networks, this can be done via ordinary
backpropagation if one considers that the activation function of the
layer representing  $Y$ is $y=f(x)+\mathcal{N}(0,\Sigma)$: one can then run
several independent samples $y_i$, backpropagate the loss obtained
with each $y_i$, and average over $i$. The backpropagation from $y$ to the
input layer can even be factorized over the samples, thanks to linearity
of backpropagation, namely: generate samples $y_i\sim
\mathcal{N}(f(x),\Sigma)$, backpropagate the error obtained with $y_i$
from the output to the layer representing $Y$, average the obtained
backpropagated values over $i$, and backpropagate from the $Y$ layer to
the input layer using $f$. For any explicit choice of $\rho$, the contribution of the gradient of the $\log \rho(y)$ term can
easily be incorporated into this scheme.

\paragraph{Optimal noise level.}
\label{sec:optnoise}
A good choice of noise level $\Sigma$ leads to tighter bounds on $\Lgen$: a small $\Sigma$
results in a high cost of encoding the features up to $\Sigma$ ($\log \det
\Sigma$ term), while a large $\Sigma$ will result in more noise on
features and a worse reconstruction error. An approximately optimal
choice of $\Sigma$ can be obtained by a Taylor expansion of the
reconstruction error around $f(x)$, as follows. (A theoretical treatment
of using such Taylor expansions for optimization with denoising can be
found in \cite{GCB97noiseinjection}.)

\begin{lem}[ (Taylor expansion of $\Lfgen$ for small $\Sigma$)]
\label{lem:taylorbound}
Let $f\from X\to Y$ be a deterministic feature function with values in
$Y=\R^d$. Let $\LNfgen$ be the upper bound \eqref{eq:gaussianbound}
using a normal distribution
$\mathcal{N}(f(x),\Sigma)$ for features. Then for small covariance
matrix $\Sigma$ we
have
\begin{equation}
\label{eq:hbound}
\LNfgen(x) \approx \Lrec(x)-\log \rho(f(x))-\frac12\log\det \Sigma
+\frac12\Tr(\Sigma H)-\frac{d}{2}(1+\log \twopi)
\end{equation}
where $\Lrec(x)$ is the deterministic reconstruction error using
feature $y=f(x)$, and $H$ is the Hessian
\begin{equation}
H=\frac{\partial^2}{\partial y^2} (\Lrec^y(x)-\log \rho(y))
\end{equation}
at $y=f(x)$, where $\Lrec^y(x)$ is the reconstruction error using feature
$y$.
Thus this is an approximate upper bound on $\Lgen(x)$.
\end{lem}

\begin{thm}[ (Optimal choice of $\Sigma$ for feature noise)]
\label{thm:optnoise}
Let $f\from X\to Y$ be a deterministic feature function with values in
$Y=\R^d$. Let $\LNfgen$ be the upper bound \eqref{eq:gaussianbound}
using a normal distribution
$\mathcal{N}(f(x),\Sigma)$ for features. Let as above
\begin{equation}
H(x)\deq\frac{\partial^2}{\partial y^2} (\Lrec^y(x)-\log \rho(y))
\end{equation}
at $y=f(x)$.

Then the choice $\Sigma(x)=H(x)^{-1}$ (provided $H$ is positive) is optimal in the bound~\eqref{eq:hbound} and yields
\begin{equation}
\label{eq:opthbound}
\LNfgen(x) \approx \Lrec(x)-\log \rho(f(x)) +\frac12\log\det
H(x) -\frac{d}{2} \log\twopi
\end{equation}
as an approximate upper bound on $\Lgen(x)$.

Among diagonal matrices $\Sigma$, the optimal choice is $\Sigma(x)=(\diag
H(x))^{-1}$ and produces a corresponding term $\frac12 \log \det \diag
H(x)$ instead of $\frac12 \log \det H(x)$.
\end{thm}

In addition to the reconstruction error $\Lrec(x)$ at $f(x)$ and
to the encoding cost $-\log \rho(f(x))$ under the elementary model, this
codelength bound involves the reconstruction error around $f(x)$ through
the Hessian.  Minimizing this bound will
favor points where the error is small in the widest possible feature
region around $f(x)$. This presumably leads to more robust
reconstruction.

Several remarks can be made on this result. First, 
the optimal choice of noise $\Sigma$ depends on the data sample $x$,
since $H$ does. This should not be a practical problem when training
denoising auto-encoders.

Second, this choice only optimizes a Taylor approximation of the actual bound
in Corollary~\ref{cor:DAE}, so it is only approximately optimal; see
\cite{GCB97noiseinjection}. Still,
Corollary~\ref{cor:DAE} applies to any choice of $\Sigma$ so it  provides
a valid, exact bound for this approximately optimal choice.

Third, computing the Hessian $H(x)$ may not be practical.  Still, since
again Corollary~\ref{cor:DAE} applies to an arbitrary $\Sigma$, it is not
necessary to compute $H(x)$ exactly, and any reasonable approximation of
$H(x)^{-1}$ yields a valid near-optimal bound and should provide a
suitable order of magnitude for feature noise.  \cite[Section 7]{LBOM96}
provides useful Hessian approximations for neural networks, in particular
the diagonal Gauss--Newton approximation (see the Appendix for more
details).

In practice there are two different ways of using this result:
\begin{itemize}
\item One can use the denoising criterion of
Corollary~\ref{cor:DAE}, in which at each step the noise level is
set to an approximation of $H(x)^{-1}$, such as diagonal Gauss--Newton. This alternates between
optimizing the model parameters for a given noise level, and optimizing
the noise level for given model parameters.
\item One can work directly with the objective function
\eqref{eq:opthbound} from Theorem~\ref{thm:optnoise}, which has an
error term $\Lrec$ and a regularization term $\log\det H(x)$. Computing a
gradient of the latter may be tricky. For multilayer neural
networks, we provide in the Appendix (Theorem~\ref{thm:gradlogH}) an
algorithm
to compute this gradient at a cost of two forward
and
backpropagation passes if the layer-wise diagonal Gauss--Newton
approximation of \cite{LBOM96} is used for $H$.
\end{itemize}

\begin{dem}[ of Lemma~\ref{lem:taylorbound}]
Using $y\sim
\mathcal{N}(f(x),\Sigma)$ in Proposition~\ref{prop:main}, the
reconstruction error $\E\Lrec(x)$ is $\E_y \Lrec^y(x)$.
Using a second-order Taylor expansion of $\Lrec^y(x)$ around $y=f(x)$, and
using that $\E_{z\sim \mathcal{N}(0,\Sigma)} (\transp{z}Mz)=\Tr(\Sigma M)$
for any matrix $M$, we find
$\E\Lrec(x)\approx \Lrec(x)+\frac12 \Tr(\Sigma H_g)$ where $H_g$ is the Hessian of
$\Lrec^y(x)$ at $y=f(x)$. 
By a similar argument the term $\KL{\mathcal{N}(f(x),\Sigma)}{\rho}$ is approximately
$-\Ent \mathcal{N}(f(x),\Sigma)-\log \rho(f(x))+\frac12 \Tr(\Sigma
H_\rho)$ with $H_\rho$ the Hessian of $-\log \rho(y)$ at $y=f(x)$.
Thus the bound $\LNfgen(x)$
is approximately $\Lrec(x)-\log \rho(f(x))-\Ent
\mathcal{N}(f(x),\Sigma)+\frac12 \Tr(\Sigma H)$ with $H=H_g+H_\rho$.
The result follows from $\Ent\mathcal{N}(f(x),\Sigma)=\frac12 \log \det \Sigma+
\frac{d}{2} (1+\log \twopi)$.
\end{dem}

\begin{dem}[ of Theorem~\ref{thm:optnoise}]
Substituting $\Sigma=H(x)^{-1}$ in~\eqref{eq:hbound} directly yields the
estimate in the proposition. Let us prove
that this choice is optimal. We have to minimize $-\log \det
\Sigma+\Tr(\Sigma H)$ over $\Sigma$. The case of diagonal $\Sigma$
follows by
direct minimization over the diagonal entries. For the general case,
we have $\Tr(\Sigma H)=\Tr(H^{1/2}\Sigma
H^{1/2})$. Since $H^{1/2}\Sigma
H^{1/2}$ is symmetric we can decompose $H^{1/2}\Sigma
H^{1/2}=\transp{O}DO$ with $O$ orthogonal and $D$ diagonal. Then
$\Tr(\Sigma H)=\Tr(\transp{O}DO)=\Tr(D)$. Moreover, $\log \det
\Sigma=\log \det(H^{-1/2}\transp{O}DOH^{-1/2})=-\log \det H+\log\det D$
so that $-\log \det \Sigma+\Tr(\Sigma H)=\log \det H-\log \det
D+\Tr(D)=\log\det H+\sum_k (d_k-\log d_k)$ with $d_k$ the entries of $D$.
The function $z\mapsto z-\log z$ is convex on $\R_+$ with a unique minimum
at $z=1$, so this is minimal if and only if $D=\Id$, i.e.,
$\Sigma=H^{-1}$.
\end{dem}

\paragraph{Link with contractive auto-encoders.}
\label{sec:contract}
The Hessian of the reconstruction error may not be easy to compute in
practice. However, when reconstruction error is small this Hessian is
related to the square derivatives of the reconstructed output with respect to the
features, using the well-known Gauss--Newton approximation.

The resulting bound on codelength penalizes large square
derivatives of the reconstructed outputs, as follows.

This is reminiscent
of contractive auto-encoders (\cite{ContractiveAE2011}; see also
\cite{Bishop95noiseregul} for the relationship between denoising and
contractivity as regularization methods), with two differences: the
contractivity is from features to output instead of from input to
features,
and instead of the Frobenius norm of the
Jacobian matrix \cite{ContractiveAE2011}, the penalty is the sum of the logs of the norms of the rows
of this matrix.


\begin{prop}[ (Codelength and contractivity)]
\label{prop:CAE}
Consider a quadratic reconstruction error of the type
$
L=\sum_k\frac{(\hat x^k-x^k)^2}{2\sigma_k^2}
$
where $\hat x^k$ are the components of the reconstructed data $\hat
x=\hat x(y)$ using features $y$. Let the elementary model $\rho$ on $Y$ be
Gaussian with variance $\diag(\lambda_i)$.

Then, when the reconstruction error is small enough,
\begin{equation}
\label{eq:contractivebound}
\Lrec(x)-\log \rho(f(x)) + \sum_i
\log \sqrt{\frac{1}{\lambda_i}+\sum_k \frac{1}{\sigma_k^2}\left(\frac{\partial \hat x^k}{\partial
y^i}\right)^2} -\frac{d}{2} \log\twopi
\end{equation}
is an approximate upper bound on $\Lgen(x)$. 
\end{prop}

This corresponds to the
approximately optimal choice
$\Sigma=(\diag H)^{-1}$ together with the Gauss--Newton approximation
$
\frac{\partial^2 L}{\partial y^i\partial y^j}\approx \sum_k \frac{1}{\sigma_k^2} \frac{\partial \hat
x^k}{\partial
y^i}\frac{\partial \hat x^k}{\partial y^j}
$.

The terms $1/\lambda_i$ prevent the logarithms from diverging to
$-\infty$ in case a
feature component $i$ has no influence on the output $\hat x$. Typically
$\lambda_i$ will be large so the Jacobian norm $\sum_k
\frac{1}{\sigma_k^2}\left(\frac{\partial \hat x^k}{\partial
y^i}\right)^2$ dominates.

\cite{ContractiveAE2011} contains an indication on how to optimize
objective functions involving such derivatives for the case of a
\emph{single-layer} neural network: in that case the square derivatives are
related to the squared weights of the network, so that the gradient
of this term can be computed. For more complex models, however,
$\partial \hat x^k/\partial y^i$ is a complex (though computable) function of the model
parameters. Computing the gradient of
$\left(\frac{\partial \hat
x^k}{\partial y^i}\right)^2$ with respect to the model parameters is thus
feasible but costly. Lemma~\ref{lem:paths} in the Appendix allows to
compute a similar quantity for multilayer networks if the Gauss--Newton
approximation is used on each layer in turn, instead of once globally
from the $y$ layer to the $\tilde x$ layer as used here. Optimizing
\eqref{eq:contractivebound}
for multilayer networks using
Lemma~\ref{lem:paths} would require $(\dim X)$ distinct
backpropagations.
More work is needed on this, such as stacking auto-encoders
\cite{HintonSalakhutdinovScience2006} to work with
only one
layer at a time.

\begin{dem}[ of Proposition~\ref{prop:CAE}]
Starting from Theorem~\ref{thm:optnoise}, we have to approximate the Hessian
$H(x)=\frac{\partial^2}{\partial y^2} (\Lrec^y(x)-\log \rho(y))$. By the assumption that
$\Lrec^y(x)=\sum_k \frac{(\hat x^k-x^k)^2}{2\sigma_k^2}$, where $\hat x$
is a function of $y$, and since $\rho$ is Gaussian, we get
\begin{equation}
H_{ij}(x)=\diag(1/\lambda_i)+\sum_k \frac{1}{2\sigma_k^2} \frac{\partial^2 }{\partial
y^i\partial y^j} (\hat
x^k(y)-x^k)^2
\end{equation}
and we can use the well-known Gauss--Newton approximation
\cite[5.4.2]{Bishop_book}, namely
\begin{align}
\frac{\partial^2 }{\partial y^i\partial y^j} (\hat
x^k(y)-x^k)^2&=2\frac{\partial \hat x^k}{\partial y^i}\frac{\partial \hat
x^k}{\partial y^j}+2\left(\hat x^k(y)-x^k\right)\frac{\partial^2 \hat x^k}{\partial
y^i\partial y^j}
\\&\approx 2\frac{\partial \hat x^k}{\partial y^i}\frac{\partial \hat
x^k}{\partial y^j}
\end{align}
valid whenever the error $\hat x^k(y)-x^k$ is small enough.
(Interestingly, when summing over the dataset, it is not necessary that
\emph{every} error is small enough, because errors with opposite signs
will compensate; a fact used implicitly in~\cite{Bishop95noiseregul}.)

The diagonal terms of $H(x)$ are thus
\begin{equation}
H_{ii}(x)\approx \frac1{\lambda_i}+\sum_k \frac{1}{\sigma_k^2}\left(\frac{\partial \hat
x^k}{\partial
y^i}\right)^2
\end{equation}

Now, from Theorem~\ref{thm:optnoise} the choice $\Sigma=(\diag
H)^{-1}$ is optimal among diagonal noise matrices $\Sigma$. Computing the
term $\frac12 \log \det \diag H=\sum_i \log \sqrt{H_{ii}}$ from
\eqref{eq:opthbound} and substituting $H_{ii}$ ends the proof.
\end{dem}

\begin{rem}
A tighter (approximately optimal) but less convenient bound is
\begin{equation}
\Lrec(x)
-\log \rho(f(x)) +\frac12
\log \det \left(-\frac{\partial^2\log \rho(y)}{\partial y^i \partial
y^j}+\sum_k \frac{1}{\sigma_k^2}\frac{\partial \hat
x^k}{\partial
y^i}\frac{\partial \hat x^k}{\partial y^j}\right)_{\!\!ij}
-\frac{d}{2} \log\twopi
\end{equation}
which forgoes the diagonal approximation.
For more general loss functions, a similar argument applies, resulting in
a more complex expression which
involves the Hessian of the loss with respect to the reconstruction $\hat
x$.
\end{rem}

\begin{rem}[ (Adapting the elementary feature model $\rho$)]
Since all our bounds on $\Lgen$ involve
$\log \rho(f(x))$ terms, the best choice of elementary model $\rho$ is the one which
maximizes the log-likelihood of the empirical feature distribution in
space $Y$. This can be done concurrently with the
optimization of the codelength, by re-adapting the prior after each step
in the optimization of the functions $f$ and $g$. For Gaussian models
$\rho$ as in Proposition~\ref{prop:CAE}, this leads to
\begin{equation}
\lambda_i\gets \Var \left[f(x)^i\right]
\end{equation}
with $f(x)^i$ the $i$-th component of feature $f(x)$, and $x$ ranging over
the dataset. If using the ``denoising'' criterion from
Corollary~\ref{cor:DAE}, the noise on $f(x)$ must be included when
computing this variance.
\end{rem}

\paragraph{Variance of the output, and relative versus absolute error.}
\label{sec:variance}
A final, important choice when considering auto-encoders from a
compression perspective is whether or not to include the variance of the
output as a model parameter. While minimizing the reconstruction error
usually focuses on absolute error,
dividing the error by two will reduce
codelength by one bit whether the error is large or small. This
works out as follows.

Consider a situation where the outputs are real-valued (e.g.,
image). The usual loss is the square loss $L=\sum_n \sum_i
(x^i_n-\hat x^i_n)^2$ where $n$ goes through all samples and $i$ goes
through the components of each sample (output dimension), 
the $x_n$ are the actual data, and the $\hat x_n$ are the reconstructed data
computed from the features $y$.

This square loss is recovered as the log-likelihood of the data over a Gaussian model with fixed
variance $\sigma$ and mean $\hat x_n$:
\begin{equation}
\Lrec(\D)=-\sum_{x\in\D} \log g_{\hat x}(x)=\sum_{x\in\D} \sum_i \left(\frac{(x^i-\hat
x^i)^2}{2\sigma^2}+\log \sigma+\frac12 \log \twopi\right)
\end{equation}
For any fixed $\sigma$, the optimum is the same as for the square loss above.

Incorporating a new parameter $\sigma_i$ for the variance of the $i$-th
component into the model may make a difference if the various output
components have different scales or noise levels. The reconstruction
error becomes
\begin{equation}
\Lrec(\D)=\sum_i \sum_{x\in\D} \left( \frac{(x^i-\hat
x^i)^2}{2\sigma_i^2}+\log \sigma_i+\frac12 \log \twopi \right)
\end{equation}
which is now to be optimized jointly over the functions $f$ and $g$, and the $\sigma_i$'s.
The optimal $\sigma_i$ for a given $f$ and $g$ is
the mean square error\footnote{If working with feature noise as in
Corollary~\ref{cor:DAE}, this is the error after adding the noise.
Optimizing $\sigma_i$ for the estimate in Proposition~\ref{prop:CAE} is
more complicated since $\sigma_i$ influences both the reconstruction
error and the regularization term.} of component $i$,
\begin{equation}
{\sigma^\ast_i}^2=E_i\deq \frac{1}{\#\D}\sum_{x\in\D} (x^i-\hat
x^i)^2
\end{equation}
so with this optimal choice the reconstruction error is
\begin{equation}
\Lrec(\D)=(\#\D)\sum_i
\left(\frac{1}{2}+\frac{1}{2}\log E_i +\frac{1}{2}\log \twopi
\right)
\end{equation}
and so we have to optimize
\begin{equation}
\Lrec(\D)=\frac{\#\D}{2}\sum_i \log E_i+\mathrm{Cst}
\end{equation}
that is, the sum of the \emph{logarithms} of the mean
square error for each component.
(Note that this is not additive over the dataset: each $E_i$ is an average
over the dataset.)
Usually, the sum of the $E_i$ themselves is used. Thus,
including the $\sigma_i$ as parameters changes the minimization problem
by focusing on \emph{relative} error,
both for codelength and reconstruction error.

This is not cancelled out by normalizing the data: indeed the above does
not depend on the variance of each component, but on the mean square
prediction error, which can vary even if all components have the same
variance, if some components are harder to predict.

This is to be used with caution when some errors become
close to $0$ (the log tends to $-\infty$). Indeed, optimizing this objective
function means that being able to predict an output component with an accuracy of $100$
digits (for every sample $x$ in the data) can balance
out $100$ bad predictions on other output components.
This is only relevant if the data are actually precise up
to $100$ significant digits.
In practice an error of $0$ only means that the actual error is below the
quantization level $\epsilon$. Thus,
numerically, we might want to consider that the smallest possible square
error is $\epsilon^2$, and to optimize $\sum_i \log (E_i+\epsilon^2)$
for data quantized up to $\epsilon$.

When working with the results of the previous sections
(Prop.~\ref{prop:main}, Corollary~\ref{cor:DAE}, Thm.~\ref{thm:optnoise},
and Prop.~\ref{prop:CAE}), changing $\sigma$ has an influence: it changes
the relative scaling of the reconstruction error term $\Lrec$ w.r.t.\ the
remaining information-theoretic terms. Choosing the optimal $\sigma_i$ as
described here fixes this problem and makes all terms homogeneous.

Intuitively, from the minimum description length or compression
viewpoint, dividing an error by $2$ is an equally good move whether the
error is small or large (one bit per sample gained on the codelength).
Still, in a specific application, the relevant loss function may be the
actual sum of square errors as usual, or a user-defined perceptual error.
But in order to find a good representation of the data as an intermediate
step in a final, user-defined problem, the compression point of view
might be preferred.

\section*{Conclusions and perspectives}

We have established that there is a strong relationship between
minimizing a codelength of the data and minimizing reconstruction error
using an auto-encoder. A variational approach provides a bound on data
codelength in terms of the reconstruction error to which certain
regularization terms are added.

The additional terms in the codelength bounds can be interpreted as a
denoising condition from features to reconstructed output. This is in
contrast with previously proposed denoising auto-encoders. For neural
networks, this criterion can be trained using standard backpropagation
techniques.

The codelength approach determines an optimal noise level for this
denoising interpretation, namely, the one that will provide the tightest
codelength. This optimal noise is approximately the inverse Hessian of
the reconstruction function, for which several approximation techniques
exist in the literature.

A practical consequence is that the noise level should be set differently
for each data sample in a denoising approach.

Under certain approximations, the codelength approach also translates as
a penalty for large derivatives from feature to output, different
from that posited in contractive auto-encoders. However, the resulting
criterion is hard to train for complex models such as multilayer neural
networks. More work is needed on this point.

Including the variances of the outputs as parameters results in better
compression bounds and a modified reconstruction error involving the
\emph{logarithms} of the square errors together with the data
quantization level. Still, having these variances as parameters is a
modeling choice that may be relevant for compression but not in
applications where the actual reconstruction error is considered.

It would be interesting to explore the practical consequences of these
insights. Another point in need of further inquiry is how this codelength
viewpoint combines with the stacking approach to deep learning, namely,
after the data $x$ have been learned using features $y$ and an elementary
model for $y$, to further learn a finer model of $y$. For instance, it is
likely that there is an interplay, in the denoising interpretation,
between the noise level used on $y$ when computing the codelength of $x$,
and the output variance $\sigma_y$  used in the definition of the
reconstruction error of a model of $y$ at the next level. This would
require modeling the transmission of noise from one layer to another in
stacked generative models and optimizing the levels of noise to minimize
a resulting bound on codelength of the output.

\bibliographystyle{alpha}
\bibliography{aagen}

\newcommand{\etalchar}[1]{$^{#1}$}
\begin{thebibliography}{RVM{\etalchar{+}}11}

\bibitem[AO12]{deeptrain}
Ludovic Arnold and Yann Ollivier.
\newblock Layer-wise training of deep generative models.
\newblock Preprint, arXiv:1212.1524, 2012.

\bibitem[Bis95]{Bishop95noiseregul}
Christopher~M. Bishop.
\newblock Training with noise is equivalent to {T}ikhonov regularization.
\newblock {\em Neural Computation}, 7(1):108--116, 1995.

\bibitem[Bis06]{Bishop_book}
Christopher~M. Bishop.
\newblock {\em Pattern recognition and machine learning}.
\newblock Springer, 2006.

\bibitem[GCB97]{GCB97noiseinjection}
Yves Grandvalet, St{\'{e}}phane Canu, and St{\'{e}}phane Boucheron.
\newblock Noise injection: Theoretical prospects.
\newblock {\em Neural Computation}, 9(5):1093--1108, 1997.

\bibitem[Gra11]{Graves_varinf}
Alex Graves.
\newblock Practical variational inference for neural networks.
\newblock In John Shawe{-}Taylor, Richard~S. Zemel, Peter~L. Bartlett, Fernando
  C.~N. Pereira, and Kilian~Q. Weinberger, editors, {\em Advances in Neural
  Information Processing Systems 24: 25th Annual Conference on Neural
  Information Processing Systems 2011. Proceedings of a meeting held 12-14
  December 2011, Granada, Spain.}, pages 2348--2356, 2011.

\bibitem[Gr{\"u}07]{GrunwaldMDL}
Peter~D. Gr{\"u}nwald.
\newblock {\em The minimum description length principle}.
\newblock MIT Press, 2007.

\bibitem[HS06]{HintonSalakhutdinovScience2006}
Geoffrey~E. Hinton and Ruslan~R. Salakhutdinov.
\newblock Reducing the dimensionality of data with neural networks.
\newblock {\em Science}, 313:504--507, 2006.

\bibitem[HvC93]{HintonCamp_varinf}
Geoffrey~E. Hinton and Drew van Camp.
\newblock Keeping the neural networks simple by minimizing the description
  length of the weights.
\newblock In Lenny Pitt, editor, {\em Proceedings of the Sixth Annual {ACM}
  Conference on Computational Learning Theory, {COLT} 1993, Santa Cruz, CA,
  USA, July 26-28, 1993.}, pages 5--13. {ACM}, 1993.

\bibitem[KW13]{VariationalAE}
Diederik~P. Kingma and Max Welling.
\newblock Stochastic gradient {VB} and the variational auto-encoder.
\newblock Preprint, arXiv:1312.6114, 2013.

\bibitem[LBOM96]{LBOM96}
Yann LeCun, L{\'e}on Bottou, Genevieve~B. Orr, and Klaus-Robert M{\"u}ller.
\newblock Efficient backprop.
\newblock In Genevieve~B. Orr and Klaus-Robert M{\"u}ller, editors, {\em Neural
  Networks: Tricks of the Trade}, volume 1524 of {\em Lecture Notes in Computer
  Science}, pages 9--50. Springer, 1996.

\bibitem[Oll13]{gradnn}
Yann Ollivier.
\newblock Riemannian metrics for neural networks {I}: feedforward networks.
\newblock Preprint, \url{http://arxiv.org/abs/1303.0818}~, 2013.

\bibitem[PH87]{HintonPlaut1987}
David~C. Plaut and Geoffrey Hinton.
\newblock Learning sets of filters using back-propagation.
\newblock {\em Computer Speech and Language}, 2:35--61, 1987.

\bibitem[RVM{\etalchar{+}}11]{ContractiveAE2011}
Salah Rifai, Pascal Vincent, Xavier Muller, Xavier Glorot, and Yoshua Bengio.
\newblock Contractive auto-encoders: Explicit invariance during feature
  extraction.
\newblock In Lise Getoor and Tobias Scheffer, editors, {\em Proceedings of the
  28th International Conference on Machine Learning, {ICML} 2011, Bellevue,
  Washington, USA, June 28 - July 2, 2011}, pages 833--840. Omnipress, 2011.

\bibitem[VLL{\etalchar{+}}10]{DenoisingAE2010}
Pascal Vincent, Hugo Larochelle, Isabelle Lajoie, Yoshua Bengio, and
  Pierre-Antoine Manzagol.
\newblock Stacked denoising autoencoders: Learning useful representations in a
  deep network with a local denoising criterion.
\newblock {\em Journal of Machine Learning Research}, 11:3371--3408, 2010.

\end{thebibliography}

\appendix
\section*{Appendix: Derivative of $\log \det H$ for multilayer neural
networks}

\small

The codelength bound from Theorem~\ref{thm:optnoise} involves a term
$\log \det H(x)$ where $H(x)$ is the Hessian of the loss function for
input $x$. Optimizing this term with respect to the model parameters is
difficult in general.

We consider the case when the generative model $g\from Y\to X$ is a
multilayer neural network. We provide an algorithm to compute the
derivative of the $\log \det H(x)$ term appearing in
Theorem~\ref{thm:optnoise} with respect to the network weights, using
the layer-wise diagonal Gauss--Newton approximation of the Hessian $H(x)$ from
\cite{LBOM96}. The algorithm has the same asymptotic computational cost as
backpropagation.

\newcommand{\actf}{s}
\newcommand{\h}{\mathfrak{h}}
\renewcommand{\L}{\mathcal{L}}
\newcommand{\Lin}{\L_\mathrm{in}}
\newcommand{\Lout}{\L_\mathrm{out}}

So let the generative model $g$ be a multilayer neural network with activation function $\actf$.
The activity of unit $i$ is
\begin{equation}
a_i\deq \actf(V_i), \qquad V_i\deq \sum_{j\to i} a_j w_{ji}
\end{equation}
where the sum includes the bias term via the always-activated unit $j=0$
with $a_j\equiv 1$.

Let $L$ be the loss function of the network.

The layer-wise diagonal Gauss--Newton approximation computes an
approximation $\h_i$ to the Hessian $\frac{\partial^2 L}{\partial a_i^2}$
in the following way \cite[Sections 7.3--7.4]{LBOM96}: On the output units $k$, $\h_k$ is
directly set to $\h_k\deq \frac{\partial^2 L}{\partial a_k^2}$, and this is
backpropagated through the network via
\begin{equation}
\label{eq:bpm}
\h_i \deq \sum_{j,\,i\to j} \left(\frac{\partial a_j}{\partial
a_i}\right)^2 \h_j =\sum_{j,\,i\to j} w_{ij}^2 \,\actf'(V_j)^2\, \h_j 
\end{equation}
so that computing $\h_i$ is similar to backpropagation using squared weights. This is
also related to the \emph{backpropagated metric} from \cite{gradnn}.

\begin{thm}[ (Gradient of the determinant of the Gauss--Newton Hessian)]
\label{thm:gradlogH}
Consider a generative model $g$ given by a multilayer neural network.
Let the reconstruction error be
$
L=\sum_k\frac{(\hat x^k-x^k)^2}{2\sigma_k^2}
$
where $\hat x^k$ are the components of the reconstructed data $\hat
x=\hat x(y)$ using features $y$. Let the elementary model $\rho$ on $Y$
be Gaussian with variance $\diag(\lambda_i)$.

Let 
$
H(x)=\frac{\partial^2}{\partial y^2} (\Lrec^y(x)-\log \rho(y))
$ as in Theorem~\ref{thm:optnoise}.
Let $\hat H(x)$ be the layer-wise diagonal Gauss--Newton approximation of $H(x)$,
namely
\begin{equation}
\hat H(x)\deq \diag\left(\lambda_i^{-1}+\h_i\right)
\end{equation}
with $\h_i$ computed from~\eqref{eq:bpm}, initialized via
$\h_k=1/\sigma_k^2$ on the output layer.

Then the derivative of $\log \det \hat H(x)$ with respect to the network
weights $w$ can be computed exactly with an algorithmic cost of two
forward and backpropagation passes.
\end{thm}

This computation is trickier than it looks because the coefficients
$\actf'(V_j)^2$ used in the backpropagation for $\h$ depend
on the weights of all units before $j$ (because $V_j$ does), not only the
units directly influencing $j$.

\newcommand{\B}{\mathfrak{B}}

\begin{dem}
Apply the following lemma with $\B=\h$, $\phi(w,V)=w^2\actf'(V)^2$, and
$\psi_i(\h_i)=\log(\lambda_i^{-1}+\h_i)$.
\end{dem}

\begin{lem}[ (Gradients of backpropagated quantities)]
\label{lem:paths}
Let $\B$ be a function of the state of a neural network computed according to the
backpropagation equation
\begin{equation}
\B_i=\sum_{j,\,i\to j} \phi_j(w_{ij},V_j)\B_j
\end{equation}
initialized with some fixed values $\B_k$ on the output layer.

Let
\begin{equation}
S\deq\sum_{i\in\Lin} \psi_i(\B_i)
\end{equation}
for some functions $\psi_i$ on the
input layer $\Lin$.

Then the derivatives of $S$ with respect to the network parameters
$w_{ij}$ can be computed at the same algorithmic cost as one forward and
two
backpropagation passes, as follows.
\begin{enumerate}
\item Compute $\B_i$ for all $i$ by backpropagation.

\item Compute the variable $\mathfrak{C}_j$ by forward propagation for all units
$j$, as
\begin{equation}
\mathfrak{C}_j\deq \sum_{i\to j} \mathfrak{C}_i \phi_j(w_{ij},V_j)
\end{equation}
initialized with $\mathfrak{C}_i=\psi'_i(\B_i)$ for $i$ in the input layer.

\item Compute the variable $D_i$ by backpropagation for all units
$i$, as
\begin{equation}
D_i\deq\sum_{k,\,k\to i} \mathfrak{C}_k \B_i \frac{\partial \phi_i(w_{ki},V_i)}{\partial
V_i}+\sum_{j,\,i\to j} \actf'(V_i) w_{ij} D_j
\end{equation}
(also used for initialization with $i$ in the output layer, with an empty sum in
the second term).
\end{enumerate}

Then the derivatives of $S$ are
\begin{equation}
\frac{\partial S}{\partial w_{ij}}=\mathfrak{C}_i \B_j \frac{\partial
\phi_j(w_{ij},V_j)}{\partial w_{ij}}
+a_i D_j
\end{equation}
for all $i,j$.
\end{lem}

Note that we assume that the values $\B_k$ used to initialize $\B$ on the
output layer are fixed (do not depend on the network weights). Any
dependency of $\B_k$ on the output layer activity values $a_k$ can,
instead, be
incorporated into $\phi_k$ via $V_k$.

\begin{dem}
We assume that the network is an arbitrary finite, directed acyclic
graph. We also assume (for simplicity only) that no unit is both an
output unit and influences other units.
We denote $i\to j$ if there is an edge from $i$ to $j$, $i>j$ if there is
a path of length $\geq 1$ from $i$ to $j$, and $i\geq j$ if $i>j$ or
$i=j$. 

The computation has a structure similar to the forward-backward
algorithm used in hidden Markov models.

For any pair of units $l$, $m$ in the network, define the
``backpropagation transfer rate'' \cite{gradnn} from $l$ to $m$ as
\begin{equation}
\tau_l^m\deq \sum_\gamma \prod_{t=1}^{\abs{\gamma}}
\phi_{\gamma_t}(w_{\gamma_{t-1}\gamma_t},V_{\gamma_t})
\end{equation}
where the sum is over all paths $\gamma$ from $l$ to $m$ in the network
(including the length-$0$ path for $l=m$),
and $\abs{\gamma}$ is the length of $\gamma$. In particular, $\tau_m^m=1$
and $\tau_l^m=0$ if there is no path from $l$ to $m$. By construction these
satisfy the backpropagation equation
\begin{equation}
\tau_i^k=\sum_{j,\,i\to j} \phi_j(w_{ij},V_j) \tau_j^k
\end{equation}
for $i\neq k$. By induction
\begin{equation}
\label{eq:Bconvol}
\B_i=\sum_{k\in \Lout} \tau_i^k \B_k
\end{equation}
where the sum is over $k$ in the output layer $\Lout$. Consequently the
derivative of $S=\sum_{i\in\Lin} \psi_i(\B_i)$ with respect to a weight $w_{mn}$ is
\begin{equation}
\label{eq:dSdwsum}
\frac{\partial S}{\partial w_{mn}}=\sum_{i\in \Lin}\psi_i'(\B_i)\sum_{k\in\Lout}
\frac{\partial \tau_i^k}{\partial w_{mn}} \B_k 
\end{equation}
so that we have to compute the derivatives of $\tau_i^k$. (This assumes
that the initialization of $\B_k$ on the output layer does not depend on
the weights $w$.)

A weight $w_{mn}$ influences $\phi_n(w_{mn},V_n)$ and also influences
$V_n$ which in turn influences all values of $V_j$ at subsequent units.
Let us first compute the derivative of $\tau_i^k$ with respect to $V_n$.
Summing over paths $\gamma$ from $i$ to $k$ we find
\begin{align}
\label{eq:dtdvpath}
\frac{\partial \tau_i^k}{\partial V_n}&=
\sum_\gamma \frac{\partial}{\partial V_n}\prod_{t=1}^{\abs{\gamma}}
\phi_{\gamma_t}(w_{\gamma_{t-1}\gamma_t},V_{\gamma_t})
\\&=\sum_\gamma \sum_t \left(\prod_{s=1}^{t-1}
\phi_{\gamma_s}(w_{\gamma_{s-1}\gamma_s},
V_{\gamma_s})\right)
\frac{\partial \phi_{\gamma_t}(w_{\gamma_{t-1}\gamma_t},V_{\gamma_t})}{\partial V_n}
\left(\prod_{s=t}^{\abs{\gamma}} \phi_{\gamma_s}(w_{\gamma_{s-1}\gamma_s},
V_{\gamma_s})\right)
\\&=
\sum_{(l,m),\,l \to m} \tau_i^l \,\frac{\partial \phi_m(w_{lm},V_m)}{\partial
V_n} \,\tau_m^k
\label{eq:dtdv}
\end{align}
by substituting $l=\gamma_{t-1}$, $m=\gamma_t$ for each value of $t$, and
unraveling the definition of $\tau_i^l$ and $\tau_m^k$.

Since $V_n$ only influences later units in the network,
the only non-zero terms are those with $n\geq m$. We can decompose into
$m=n$ and $n>m$:
\begin{align}
\frac{\partial \tau_i^k}{\partial V_n}
&=
\sum_{l,\,l \to n} \tau_i^l \frac{\partial \phi_n(w_{ln},V_n)}{\partial
V_n} \tau_n^k
+
\sum_{m,n>m}\sum_{l,\,l \to m} \tau_i^l \frac{\partial \phi_m(w_{lm},V_m)}{\partial
V_n} \tau_m^k
\label{eq:dtdvdecomp}
\end{align}

Now, for $n>m$, the influence of $V_n$ on $V_m$ has to transit through
some unit $j$ directly connected to $n$, namely, for any function
$\mathcal{F}(V_m)$,
\begin{equation}
\frac{\partial \mathcal{F}(V_m)}{\partial V_n}=\sum_{j,\, n\to j} 
\actf'(V_n)w_{nj}
\frac{\partial \mathcal{F}(V_m)}{\partial V_j}
\end{equation}
where $\actf$ is the activation function of the network.
So
\begin{align}
\sum_{m,\,n>m} \sum_{l,\,l\to m} \tau_i^l 
\frac{\partial \phi_m(w_{lm},V_m)}{\partial
V_n} \tau_m^k
&=\sum_{j,\,n\to j} \actf'(V_n)w_{nj} \sum_{m,\,n>m} \sum_{l,\,l\to m} \tau_i^l 
\frac{\partial \phi_m(w_{lm},V_m)}{\partial
V_j} \tau_m^k
\\&=\sum_{j,\,n\to j} \actf'(V_n)w_{nj} \sum_{m} \sum_{l,\,l\to m} \tau_i^l 
\frac{\partial \phi_m(w_{lm},V_m)}{\partial
V_j} \tau_m^k
\label{eq:dtdvbp}
\end{align}
where the difference between the last two lines is that we
removed the condition $n>m$
in the summation over $m$: indeed, any $m$ with non-vanishing
$\partial V_m/\partial V_j$ satisfies $j\geq m$ hence $n>m$.
According to \eqref{eq:dtdv}, $\sum_m \sum_{l,\,l\to m} \tau_i^l
\frac{\partial \phi_m(w_{lm},V_m)}{\partial
V_j} \tau_m^k$ is $\frac{\partial \tau_i^k}{\partial V_j}$, so that
\eqref{eq:dtdvbp}
is $\sum_{j,\,n\to j} \actf'(V_n)w_{nj} \frac{\partial
\tau_i^k}{\partial V_j}$.
Collecting from \eqref{eq:dtdvdecomp}, we find
\begin{equation}
\label{eq:bpdtdv}
\frac{\partial \tau_i^k}{\partial V_n}=
\sum_{l,\,l\to n} \tau_i^l \frac{\partial
\phi_n(w_{ln},V_n)}{\partial
V_n} \tau_n^k +
\sum_{j,\,n\to j} \actf'(V_n)w_{nj} \frac{\partial
\tau_i^k}{\partial V_j}
\end{equation}
so that the quantities $\frac{\partial \tau_i^k}{\partial V_n}$ can be
computed by backpropagation on $n$, if the $\tau$ are known.

To compute the derivatives of $\tau_i^k$ with respect to a weight
$w_{mn}$, observe that $w_{mn}$ influences the $w_{mn}$ term in
$\phi_n(w_{mn},V_n)$, as well as all terms $V_l$ with $n\geq l$ via its
influence on $V_n$. Since $\frac{\partial V_n}{\partial
w_{mn}}=a_m$ we find
\begin{equation}
\frac{\partial \phi_l(w_{jl},V_l)}{\partial w_{mn}}
=\1_{(j,l)=(m,n)} \frac{\partial \phi_n(w_{mn},V_n)}{\partial w_{mn}}
+ a_m\frac{\partial
\phi_n(w_{jl},V_l)}{\partial V_n}
\end{equation}
By following the same procedure as in \eqref{eq:dtdvpath}--\eqref{eq:dtdv} we obtain
\begin{align}
\frac{\partial \tau_i^k}{\partial w_{mn}}&=
\tau_i^m \frac{\partial \phi_n(w_{mn},V_n)}{\partial w_{mn}}
\tau_n^k+a_m \sum_{(j,l),\,j\to l}
\tau_i^j\, \frac{\partial \phi_l(w_{jl},V_l)}{\partial V_n}
\tau_l^k
\\&=
\tau_i^m \frac{\partial \phi_n(w_{mn},V_n)}{\partial w_{mn}}
\tau_n^k+a_m \frac{\partial \tau_i^k}{\partial V_n}
\label{eq:dtdw}
\end{align}
by \eqref{eq:dtdvpath}.

This allows, in principle, to compute the desired derivatives. 
By~\eqref{eq:dSdwsum} we have to compute the sum of~\eqref{eq:dtdw} over $i\in \Lin$ and $k\in\Lout$ weighted by
$\psi_i'(\B_i)$ and
 $\B_k$. This avoids a
full computation of all transfer rates $\tau$ and yields
\begin{equation}
\frac{\partial S}{\partial w_{mn}}=\mathfrak{C}_m \frac{\partial
\phi_n(w_{mn},V_n)}{\partial w_{mn}}\,\B_n 
+a_m D_n
\end{equation}
where we have set
\begin{equation}
\mathfrak{C}_m\deq \sum_{i\in \Lin} \psi_i'(\B_i)\tau_i^m,
\end{equation}
and
\begin{equation}
D_n\deq \sum_{i\in\Lin} \sum_{k\in\Lout} \psi'_i(\B_i) \B_k
\frac{\partial \tau_i^k}{\partial V_n}
\end{equation}
and where we have used that $\B$ satisfies
\begin{equation}
\B_m= \sum_{k\in\Lout} \tau_m^k \B_k
\end{equation}
by \eqref{eq:Bconvol}.

It remains to provide ways to compute
$\mathfrak{C}_m$ and
$D_n$. For $\mathfrak{C}_m$, note that the transfer rates $\tau$ satisfy
the forward propagation equation
\begin{equation}
\tau_i^k=\sum_{j,\,j\to k} \phi_k(w_{jk},V_k) \tau_i^j
\end{equation}
by construction. Summing over $i\in\Lin$ with weights $\psi'_i(\B_i)$
yields the forward propagation equation for $\mathfrak{C}$ given in the
statement of the lemma.

Finally, by summing over $i$ and $k$ in \eqref{eq:bpdtdv}, with weights
$\psi'_i(\B_i)\B_k$, and using the definition of $\mathfrak{C}$ and again the
property $\B_n=\sum_{k\in\Lout} \tau_n^k\B_k$, we obtain
\begin{equation}
D_n=\sum_{l,\,l\to n} \mathfrak{C}_l \frac{\partial \phi_n(w_{ln},V_n)}{\partial
V_n}\,\B_n+\sum_{j,\,n\to j} \actf'(V_n) w_{nj} D_j
\end{equation}
which is the backpropagation equation for $D_n$ and concludes the proof.
\end{dem}

\end{document}